\documentclass[10pt,twocolumn,letterpaper]{article}

\usepackage{iccv}
\usepackage{times}
\usepackage{epsfig}
\usepackage{graphicx}
\usepackage{amsmath}
\usepackage{amssymb}

\usepackage{nicefrac}
\usepackage{enumitem}
\usepackage{multirow}
\usepackage[outline]{contour}

\usepackage{mathrsfs}
\usepackage{scalerel}
\usepackage{diagbox}

\usepackage{tikz}
\DeclareMathAlphabet{\mathpzc}{OT1}{pzc}{m}{it}

\usepackage{tikz}
\usetikzlibrary{arrows,shapes,calc,matrix,fit,backgrounds}
\usepackage{pgfplots}
\usepackage{pgfplotstable}
\pgfplotsset{compat=1.9}

\usepackage{xstring}

\usepgfplotslibrary{external}

\newcommand{\extdata}[1]{\input{#1}}
\IfBeginWith*{\jobname}{fig/extern/}{\finalcopy}{}

\newcommand{\leg}[1]{\addlegendentry{#1}}

\tikzset{every mark/.append style={solid}}
\pgfplotsset{
	grid=both, width=\columnwidth, try min ticks=5,
	every axis/.append style={font=\scriptsize},
	every axis plot/.append style={thick,mark=none,mark size=1.2,tension=0.18},
	legend cell align=left, legend style={fill opacity=0.8},
}

\pgfplotsset{
	dash/.style={mark=o,dashed,opacity=0.7},
	dott/.style={mark=o,dotted,opacity=0.7},
}

\usepackage[pagebackref=true,breaklinks=true,letterpaper=true,colorlinks,bookmarks=false]{hyperref}

\iccvfinalcopy 

\def\iccvPaperID{5008} 

\ificcvfinal\fi
\begin{document}

\title{Targeted Mismatch Adversarial Attack:\\ {\large Query with a Flower to Retrieve the Tower}}

\author{
Giorgos Tolias \qquad Filip Radenovic \qquad Ond{\v r}ej Chum\\
Visual Recognition Group, Faculty of Electrical Engineering, Czech Technical University in Prague\\
}

\maketitle

\newcommand{\ppir}{PPIR\xspace}

\renewcommand{\paragraph}[1]{\vspace{.3\baselineskip}\noindent{\bf #1}\xspace}

\def\roxf{$\mathcal{R}$Oxford\xspace}
\def\rox{$\mathcal{R}$Oxf\xspace}
\def\ro{$\mathcal{R}$O\xspace}
\def\rpar{$\mathcal{R}$Paris\xspace}
\def\rpa{$\mathcal{R}$Par\xspace}
\def\rp{$\mathcal{R}$P\xspace}
\def\r1m{$\mathcal{R}$1M\xspace}
\def\rs{$\mathcal{R}$100k\xspace}

\newcommand{\alexnet}{\ensuremath{\mathpzc{A}}}
\newcommand{\resnetsmall}{\ensuremath{\mathpzc{R}}}
\newcommand{\vgg}{\ensuremath{\mathpzc{V}}}
\newcommand{\ensemble}{\ensuremath{\mathpzc{E}}}

\newcommand{\adve}[3]{\ensuremath{(#1{,}#2{,}#3)}}
\newcommand{\test}[3]{\ensuremath{[#1{,}\text{#2}{,}#3]}}
\newcommand{\trans}[2]{\ensuremath{#1{\rightarrow}#2}}

\newcommand{\scaleorg}{\ensuremath{{\cS_0}}\xspace}
\newcommand{\scalessparse}{\ensuremath{{\cS_1}}\xspace}
\newcommand{\scalesdense}{\ensuremath{{\cS_2}}\xspace}
\newcommand{\scalesdenselog}{\ensuremath{{\cS_3}}\xspace}

\newcommand{\hscaleorg}{\ensuremath{{\hat{\cS}_0}}\xspace}
\newcommand{\hscalessparse}{\ensuremath{{\hat{\cS}_1}}\xspace}
\newcommand{\hscalesdense}{\ensuremath{{\hat{\cS}_2}}\xspace}
\newcommand{\hscalesdenselog}{\ensuremath{{\hat{\cS}_3}}\xspace}

\newcommand{\ldesc}[1]{\ensuremath{L_{\scaleto{\text{#1}}{1.0ex}}}\xspace}
\newcommand{\lpall}{\ensuremath{L_{\scaleto{\cP}{1.0ex}}}\xspace}
\newcommand{\lhist}{\ensuremath{L_{\scaleto{\text{hist}}{1.0ex}}}\xspace}
\newcommand{\ltens}{\ensuremath{L_{\scaleto{\text{tens}}{1.0ex}}}\xspace}

\newcommand{\ldescms}[2]{\ensuremath{L_{\scaleto{\text{#1}}{1.0ex}}^{\scaleto{#2}{1.5ex}}}\xspace}
\newcommand{\ldesccms}[2]{\ensuremath{L_{\scaleto{#1}{1.0ex}}^{\scaleto{#2}{1.5ex}}}\xspace}
\newcommand{\lpallms}[1]{\ensuremath{L_{\cP}^{\scaleto{#1}{1.5ex}}}\xspace}
\newcommand{\lhistms}[1]{\ensuremath{L_{\scaleto{\text{hist}}{1.0ex}}^{\scaleto{#1}{1.5ex}}}\xspace}
\newcommand{\ltensms}[1]{\ensuremath{L_{\scaleto{\text{tens}}{1.0ex}}^{\scaleto{#1}{1.5ex}}}\xspace}

\def\lone{\ensuremath{l_1}\xspace}
\def\ltwo{\ensuremath{l_2}\xspace}
\def\linf{\ensuremath{l_\infty}\xspace}

\newcommand{\comment} [1]{{\color{orange} \Comment     #1}} 

\def\nlsp{\hspace{-12pt}}
\def\nmsp{\hspace{-5pt}}
\def\nssp{\hspace{-3pt}}
\def\nxssp{\hspace{-1pt}}
\def\xssp{\hspace{1pt}}
\def\ssp{\hspace{3pt}}
\def\msp{\hspace{5pt}}
\def\lsp{\hspace{12pt}}


\newcommand{\head}[1]{{\smallskip\noindent\bf #1}}
\newcommand{\alert}[1]{{\color{red}{#1}}}
\newcommand{\remove}[1]{{\color{blue}{#1}}}
\newcommand{\equ}[1]{(\ref{equ:#1})\xspace}

\newcommand{\red}[1]{{\color{red}{#1}}}
\newcommand{\blue}[1]{{\color{blue}{#1}}}
\newcommand{\green}[1]{{\color{green}{#1}}}
\newcommand{\gray}[1]{{\color{gray}{#1}}}


\newcommand{\tran}{^\top}
\newcommand{\mtran}{^{-\top}}
\newcommand{\zcol}{\mathbf{0}}
\newcommand{\zrow}{\zcol\tran}

\newcommand{\ind}{\mathbbm{1}}
\newcommand{\expect}{\mathbb{E}}
\newcommand{\nat}{\mathbb{N}}
\newcommand{\zahl}{\mathbb{Z}}
\newcommand{\real}{\mathbb{R}}
\newcommand{\proj}{\mathbb{P}}
\newcommand{\prob}{\mathbf{Pr}}

\newcommand{\mif}{\textrm{if }}
\newcommand{\other}{\textrm{otherwise}}
\newcommand{\minimize}{\textrm{minimize }}
\newcommand{\maximize}{\textrm{maximize }}
\newcommand{\st}{\textrm{subject to }}

\newcommand{\id}{\operatorname{id}}
\newcommand{\const}{\operatorname{const}}
\newcommand{\sgn}{\operatorname{sgn}}
\newcommand{\var}{\operatorname{Var}}
\newcommand{\mean}{\operatorname{mean}}
\newcommand{\trace}{\operatorname{tr}}
\newcommand{\diag}{\operatorname{diag}}
\newcommand{\vect}{\operatorname{vec}}
\newcommand{\cov}{\operatorname{cov}}

\newcommand{\softmax}{\operatorname{softmax}}
\newcommand{\clip}{\operatorname{clip}}

\newcommand{\defn}{\mathrel{:=}}
\newcommand{\peq}{\mathrel{+\!=}}
\newcommand{\meq}{\mathrel{-\!=}}

\newcommand{\floor}[1]{\left\lfloor{#1}\right\rfloor}
\newcommand{\ceil}[1]{\left\lceil{#1}\right\rceil}
\newcommand{\inner}[1]{\left\langle{#1}\right\rangle}
\newcommand{\norm}[1]{\left\|{#1}\right\|}
\newcommand{\frob}[1]{\norm{#1}_F}
\newcommand{\card}[1]{\left|{#1}\right|\xspace}
\newcommand{\diff}{\mathrm{d}}
\newcommand{\der}[3][]{\frac{d^{#1}#2}{d#3^{#1}}}
\newcommand{\pder}[3][]{\frac{\partial^{#1}{#2}}{\partial{#3^{#1}}}}
\newcommand{\ipder}[3][]{\partial^{#1}{#2}/\partial{#3^{#1}}}
\newcommand{\dder}[3]{\frac{\partial^2{#1}}{\partial{#2}\partial{#3}}}

\newcommand{\wb}[1]{\overline{#1}}
\newcommand{\wt}[1]{\widetilde{#1}}

\def\xssp{\hspace{1pt}}
\def\ssp{\hspace{3pt}}
\def\msp{\hspace{5pt}}
\def\lsp{\hspace{12pt}}

\newcommand{\cA}{\mathcal{A}}
\newcommand{\cB}{\mathcal{B}}
\newcommand{\cC}{\mathcal{C}}
\newcommand{\cD}{\mathcal{D}}
\newcommand{\cE}{\mathcal{E}}
\newcommand{\cF}{\mathcal{F}}
\newcommand{\cG}{\mathcal{G}}
\newcommand{\cH}{\mathcal{H}}
\newcommand{\cI}{\mathcal{I}}
\newcommand{\cJ}{\mathcal{J}}
\newcommand{\cK}{\mathcal{K}}
\newcommand{\cL}{\mathcal{L}}
\newcommand{\cM}{\mathcal{M}}
\newcommand{\cN}{\mathcal{N}}
\newcommand{\cO}{\mathcal{O}}
\newcommand{\cP}{\mathcal{P}}
\newcommand{\cQ}{\mathcal{Q}}
\newcommand{\cR}{\mathcal{R}}
\newcommand{\cS}{\mathcal{S}}
\newcommand{\cT}{\mathcal{T}}
\newcommand{\cU}{\mathcal{U}}
\newcommand{\cV}{\mathcal{V}}
\newcommand{\cW}{\mathcal{W}}
\newcommand{\cX}{\mathcal{X}}
\newcommand{\cY}{\mathcal{Y}}
\newcommand{\cZ}{\mathcal{Z}}

\newcommand{\vA}{\mathbf{A}}
\newcommand{\vB}{\mathbf{B}}
\newcommand{\vC}{\mathbf{C}}
\newcommand{\vD}{\mathbf{D}}
\newcommand{\vE}{\mathbf{E}}
\newcommand{\vF}{\mathbf{F}}
\newcommand{\vG}{\mathbf{G}}
\newcommand{\vH}{\mathbf{H}}
\newcommand{\vI}{\mathbf{I}}
\newcommand{\vJ}{\mathbf{J}}
\newcommand{\vK}{\mathbf{K}}
\newcommand{\vL}{\mathbf{L}}
\newcommand{\vM}{\mathbf{M}}
\newcommand{\vN}{\mathbf{N}}
\newcommand{\vO}{\mathbf{O}}
\newcommand{\vP}{\mathbf{P}}
\newcommand{\vQ}{\mathbf{Q}}
\newcommand{\vR}{\mathbf{R}}
\newcommand{\vS}{\mathbf{S}}
\newcommand{\vT}{\mathbf{T}}
\newcommand{\vU}{\mathbf{U}}
\newcommand{\vV}{\mathbf{V}}
\newcommand{\vW}{\mathbf{W}}
\newcommand{\vX}{\mathbf{X}}
\newcommand{\vY}{\mathbf{Y}}
\newcommand{\vZ}{\mathbf{Z}}

\newcommand{\va}{\mathbf{a}}
\newcommand{\vb}{\mathbf{b}}
\newcommand{\vc}{\mathbf{c}}
\newcommand{\vd}{\mathbf{d}}
\newcommand{\ve}{\mathbf{e}}
\newcommand{\vf}{\mathbf{f}}
\newcommand{\vg}{\mathbf{g}}
\newcommand{\vh}{\mathbf{h}}
\newcommand{\vi}{\mathbf{i}}
\newcommand{\vj}{\mathbf{j}}
\newcommand{\vk}{\mathbf{k}}
\newcommand{\vl}{\mathbf{l}}
\newcommand{\vm}{\mathbf{m}}
\newcommand{\vn}{\mathbf{n}}
\newcommand{\vo}{\mathbf{o}}
\newcommand{\vp}{\mathbf{p}}
\newcommand{\vq}{\mathbf{q}}
\newcommand{\vr}{\mathbf{r}}
\newcommand{\Vs}{\mathbf{s}}
\newcommand{\vt}{\mathbf{t}}
\newcommand{\vu}{\mathbf{u}}
\newcommand{\vv}{\mathbf{v}}
\newcommand{\vw}{\mathbf{w}}
\newcommand{\vx}{\mathbf{x}}
\newcommand{\vy}{\mathbf{y}}
\newcommand{\vz}{\mathbf{z}}

\newcommand{\vone}{\mathbf{1}}
\newcommand{\vzero}{\mathbf{0}}

\newcommand{\valpha}{{\boldsymbol{\alpha}}}
\newcommand{\vbeta}{{\boldsymbol{\beta}}}
\newcommand{\vgamma}{{\boldsymbol{\gamma}}}
\newcommand{\vdelta}{{\boldsymbol{\delta}}}
\newcommand{\vepsilon}{{\boldsymbol{\epsilon}}}
\newcommand{\vzeta}{{\boldsymbol{\zeta}}}
\newcommand{\veta}{{\boldsymbol{\eta}}}
\newcommand{\vtheta}{{\boldsymbol{\theta}}}
\newcommand{\viota}{{\boldsymbol{\iota}}}
\newcommand{\vkappa}{{\boldsymbol{\kappa}}}
\newcommand{\vlambda}{{\boldsymbol{\lambda}}}
\newcommand{\vmu}{{\boldsymbol{\mu}}}
\newcommand{\vnu}{{\boldsymbol{\nu}}}
\newcommand{\vxi}{{\boldsymbol{\xi}}}
\newcommand{\vomikron}{{\boldsymbol{\omikron}}}
\newcommand{\vpi}{{\boldsymbol{\pi}}}
\newcommand{\vrho}{{\boldsymbol{\rho}}}
\newcommand{\vsigma}{{\boldsymbol{\sigma}}}
\newcommand{\vtau}{{\boldsymbol{\tau}}}
\newcommand{\vupsilon}{{\boldsymbol{\upsilon}}}
\newcommand{\vphi}{{\boldsymbol{\phi}}}
\newcommand{\vchi}{{\boldsymbol{\chi}}}
\newcommand{\vpsi}{{\boldsymbol{\psi}}}
\newcommand{\vomega}{{\boldsymbol{\omega}}}

\newcommand{\rLambda}{\mathrm{\Lambda}}
\newcommand{\rSigma}{\mathrm{\Sigma}}

\makeatletter
\DeclareRobustCommand\onedot{\futurelet\@let@token\@onedot}
\def\@onedot{\ifx\@let@token.\else.\null\fi\xspace}
\def\eg{\emph{e.g}\onedot} \def\Eg{\emph{E.g}\onedot}
\def\ie{\emph{i.e}\onedot} \def\Ie{\emph{I.e}\onedot}
\def\cf{\emph{cf}\onedot} \def\Cf{\emph{C.f}\onedot}
\def\etc{\emph{etc}\onedot} \def\vs{\emph{vs}\onedot}
\def\wrt{w.r.t\onedot} \def\dof{d.o.f\onedot}
\def\etal{\emph{et al}\onedot}
\makeatother

\begin{abstract}
Access to online visual search engines implies sharing of private user content -- the query images.
We introduce the concept of targeted mismatch attack for deep learning based retrieval systems to generate an adversarial image to conceal the query image. The generated image looks nothing like the user intended query, but leads to identical or very similar retrieval results.
Transferring attacks to fully unseen networks is challenging. We show successful attacks to partially unknown systems, by designing various loss functions for the adversarial image construction. 
These include loss functions, for example, for unknown global pooling operation or unknown input resolution by the retrieval system.
We evaluate the attacks on standard retrieval benchmarks and compare the results retrieved with the original and adversarial image.
\end{abstract}

\vspace{-5pt}
\section{Introduction}
\label{sec:intro}
\vspace{-5pt}

\begin{figure}[t]
\vspace{-4pt}
\input{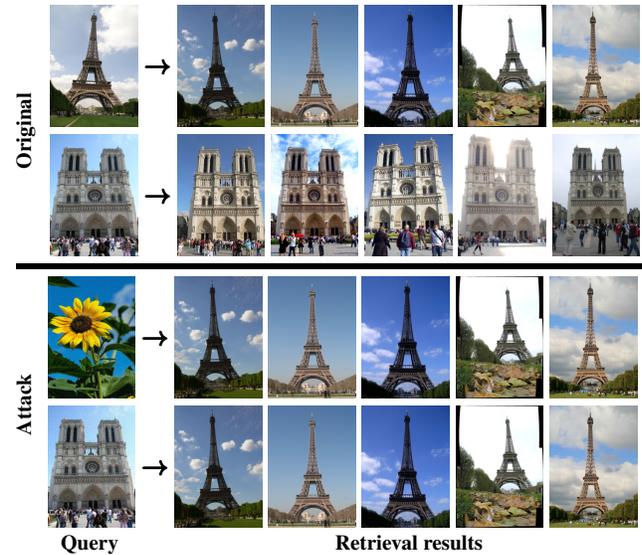}
\caption{Top two rows show retrieval results to the user query image (target). Bottom two rows show the results of our attack where a carrier image (flower, Notre Dame) is perturbed to have identical descriptor to that of the target in the first row. Identical results are obtained without disclosing the target.
\label{fig:intro_teaser}
\vspace{-12pt}
}
\end{figure}

Information about users is a valuable article. Websites, service providers, and even operating systems collect and store user data.
The collected data have various forms, \eg visited websites, interactions between users in social networks, hardware fingerprints, keyboard typing or mouse movement patterns, \etc.
Internet search engines record what the users search for, as well as the responses, \ie clicks, to the returned results. 

Recent development in computer vision allowed efficient and precise large scale image search engines to be launched, such as Google Image Search.
Nevertheless, similarly to text search engines, queries -- the images -- are stored and further analyzed by the provider\footnote{\scriptsize Google Search Help: ``The pictures you upload in your search may be stored by Google for 7 days. They won't be a part of your search history, and we'll only use them during that time to make our products and services better.''}. In this work, we protect the user  image  ({\em target}) by constructing a novel image.  The constructed image is visually dissimilar to the target, however, when used as a query, identical results are retrieved as with the target image. 
Large-scale search methods require short-code image representation, both for storage minimization and for search efficiency, which are usually extracted with Convolutional Neural Networks (CNN). We formulate the problem as an adversarial attack on CNNs.

Adversarial attacks, as introduced by Szegedy~\etal~\cite{SZS+14}, study \emph{imperceptible} non-random image perturbations to mislead a neural network.
The first attacks were introduced and tested on image classification. 
In that context, adversarial attacks are divided into two categories, namely \emph{non-targeted} and \emph{targeted}. 
The goal of non-targeted attacks is to change the prediction of a test image to an arbitrary class~\cite{MFF16, MFFF17}, while targeted attacks attempt to make a specific change of the network prediction, \ie to misclassify the test image to a predefined target class~\cite{SZS+14,CW17,DLP+18}.

Similarly to image classification, adversarial attacks have been proposed in the domain of image retrieval too.
An non-targeted attack attempts to generate an image that for a human observer carries the same visual information, while for the neural network it appears dissimilar to other images of the same object~\cite{LJL+18,LZL19,ZZH+18}. 
This way, a user protects personal images and does not allow them to be indexed for content-based search, even when the images are publicly available. 
In this paper, we address targeted attacks aiming to retrieve images that are related to a hidden target query without explicitly revealing the image (see Figure~\ref{fig:intro_teaser}).
Example applications include users checking whether their copyrighted image, or personal photo with sensitive content, \etc is indexed, \ie
used by anyone else, without providing the query image itself. Such case of privacy protection is an example of ``legal'' motivation.	
A concept that bears resemblance to ours exists in the speech recognition, but in a malicious context. 
Carlini \etal~\cite{CMV+16} generate \emph{hidden voice commands} that are imperceivable to human listeners but are interpreted as commands by devices.
We investigate adversarial attacks beyond the white-box scenario, in which all the parameters and design choices of the retrieval system are known. Specifically, we analyze the cases of unknown indexing image resolution and unknown global pooling used in the network.

\section{Related work}
\label{sec:related}

\paragraph{Adversarial attacks on image classification} were introduced by Szegedy~\etal~\cite{SZS+14}.
Follow up approaches are categorized to \emph{white-box} attacks~\cite{SZS+14,GSS15} if there is complete knowledge of the model or to \emph{black-box}~\cite{PMG16,PMG+17} otherwise.
Adversarial images are generated by various methods in the literature, such as optimization-based approaches using box-constrained L-BFGS optimizer~\cite{SZS+14}, gradient descent with change of variable~\cite{CW17}.
A fast gradient sign method~\cite{GSS15} and variants~\cite{KGB17,DLP+18} are designed to be \emph{fast} rather than optimal, while DeepFool~\cite{MFF16} analytically derives an optimal solution method by assuming that neural networks are totally linear.
All these approaches solve an optimization problem given a test image and its associated class in the case of non-targeted attacks or a test image and a target class in the case of targeted attacks.
A universal non-targeted approach is proposed by Moosavi~\etal~\cite{MFFF17}, where an image-agnostic Universal Adversarial Perturbation (UAP) is computed and applied to unseen images to cause network misclassification. 

\paragraph{Adversarial attacks on image retrieval}
are studied by recent work~\cite{LJL+18,LZL19,ZZH+18} in a non-targeted scenario for CNN-based approaches.
Liu \etal~\cite{LZL19} and Zheng \etal~\cite{ZZH+18} adopt the optimization-based approach~\cite{SZS+14}, while Li \etal~\cite{LJL+18} adopt the UAP~\cite{MFFF17}.
Similar attacks on classical retrieval systems that are based on SIFT local descriptors~\cite{L04} have been addressed in an earlier line of work by Do~\etal~\cite{DKFA10,DKAF12}.
To the best of our knowledge, no existing work focuses on targeted adversarial attacks for image retrieval.
Targeted attacks for nearest neighbors in high dimensional spaces are studied by Amsaleg \etal~\cite{ABB+17}, where they directly perturb the high dimensional vectors and show that the high local intrinsic dimensionality results in high vulnerability.

\section{Background}
\label{sec:background}
We provide the background for non-targeted and targeted adversarial attacks in the domain of image classification, then detail the basic components of CNN-based image retrieval approaches, and finally discuss non-targeted attacks for image retrieval.
All variants presented in this section assume white-box access to the network classifier for classification or the feature extractor network for retrieval.

\subsection{Image classification attacks}
\label{sec:classattack}
We denote the initial RGB image, called the \emph{carrier image}, by tensor $\vx_c \in [0,1]^{W \times H \times 3}$, and its associated label by $y_c \in \{1\ldots K\}$.
A CNN trained for $K$-way classification, denoted by function $f: \real^{W \times H \times 3} \rightarrow \real^{K}$, produces vector $f(\vx_c)$ comprising class confidence values.
Adversarial attack methods for classification typically study the case of images with correct class prediction, \ie $\arg\max_i f(\vx_c)_i$ is equal to $y_c$, where $f(\vx_c)_i$ is the $i$-th dimension of vector $f(\vx_c)$.
An adversary aims at generating \emph{adversarial image} $\vx_a$ that is visually similar to the carrier image but is classified incorrectly by $f$.
The goal of the attack can vary~\cite{AM18} and corresponds to different loss functions optimizing $\vx \in [0,1]^{W\times H \times 3}$.

\paragraph{Non-targeted misclassification} is achieved by reducing the confidence for class $y_c$, while increasing for all other classes. It is achieved by minimizing loss function
\begin{equation}
L_{\text{nc}}(\vx_c, y_c; \vx) = -\ell_{\text{ce}} (f(\vx), y_c) + \lambda~||\vx - \vx_c||^2.
\label{equ:nontargetclass}
\end{equation}
Function $\ell_{\text{ce}} (f(\vx), y_c)$ is the cross-entropy loss which is maximized to achieve misclassification. In this way, misclassification is performed to any wrong class. 
Term $||\vx - \vx_c||^2$ is called \emph{carrier distortion} or simply \emph{distortion} and is the squared \ltwo norm of the perturbation vector $\vr = \vx - \vx_c$. Other norms, such as \linf, are also applicable~\cite{CW17}.

\paragraph{Targeted misclassification} has the goal of generating an adversarial image that gets classified into target class $y_t$. It is achieved by minimizing loss function
\begin{equation}
L_{\text{tc}}(\vx_c, y_t; \vx) = \ell_{\text{ce}} (f(\vx), y_t) + \lambda~||\vx - \vx_c||^2.
\label{equ:targetclass}
\end{equation}
In contrast to \equ{nontargetclass}, cross-entropy loss is minimized \wrt the target class instead of maximized \wrt the carrier class.

\paragraph{Optimization} of \equ{nontargetclass} or \equ{targetclass} generates the adversarial images given by
\begin{equation}
\vx_a = \arg\min_{\vx} L_{\text{nc}}(\vx_c, y_c; \vx),
\end{equation}
or 
\begin{equation}
\vx_a = \arg\min_{\vx} L_{\text{tc}}(\vx_c, y_t; \vx),
\end{equation}
respectively.
In the literature~\cite{SZS+14,CW17}, various optimizers such as Adam~\cite{KB15}, or L-BFGS~\cite{BLN+95} are used.
The box constraints, \ie ~$\vx \in [0,1]^{W\times H\times 3}$, are ensured by projected gradient descent, clipped gradient descent, change of variables~\cite{CW17}, or optimization algorithms that support box constraints such as L-BFGS. It is a common practice to perform line search for weight $\lambda > 0$ and keep the attack of minimum distortion. 
The optimization is initialized by the carrier image.

\subsection{Image retrieval components}
\label{sec:ircomp}
This work focuses on attacks on CNN-based image retrieval with global image descriptors.
An image is mapped to a high dimensional descriptor by a CNN with a global pooling layer.
The descriptor is consequently normalized to have unit $\ltwo$ norm.
Then, retrieval from a large dataset \wrt a \emph{query image} reduces to nearest neighbor search via inner product evaluation between the query descriptor and dataset descriptors. 
The model for descriptor extraction consists of the following components or parameters.
\begin{itemize}[label={},leftmargin=2pt]
\setlength\itemsep{2pt}
\item \textit{Image resolution:} The input image $\vx$ is re-sampled to image $\vx^s$ to have the largest dimension equal to $s$.
\item \textit{Feature extraction:} Image $\vx^s$ is fed as an input to a Fully Convolutional Network (FCN), denoted by function $g: \real^{W \times H \times 3} \rightarrow \real^{w \times h \times d}$, which maps $\vx^s$ to tensor $g(\vx^s)$. When the image is fed at its original resolution we denote it by $g(\vx)$.
\item \textit{Pooling:} A global pooling operation $h: \real^{w \times h \times d} \rightarrow \real^d $ maps the input tensor $g(\vx^s)$ to descriptor $(h \circ g)(\vx^s)$. We assume that \ltwo normalization is included in this process, so that the output descriptor has unit \ltwo norm. 
We consider various options for pooling, namely, max pooling (MAC)~\cite{RSMC14,TSJ16}, sum pooling (SPoC)~\cite{BL15}, generalized mean pooling (GeM)~\cite{RTC18}, regional max pooling (R-MAC)~\cite{TSJ16}, and spatially and channel-wise weighted sum pooling (CroW)~\cite{KMO16}.
The framework can be extended to multiple other variants~\cite{OHB16,MMOS+16,AGTPS16}.
\item \textit{Whitening:} Descriptor post-processing is performed by function $w: \real^d \rightarrow \real^d$, which includes centering, whitening and \ltwo re-normalization~\cite{RTC18}. Finally, input image $\vx^s$ is mapped to descriptor $(w \circ h \circ g)(\vx^s)$.
\end{itemize}

For brevity we denote $\vg_\vx = g(\vx)$, $\vh_\vx = (h\circ g)(\vx)$, and $\vw_\vx = (w\circ h\circ g)(\vx)$. 
In the following, we consider an extraction model during the adversarial image optimization and another one during the testing of the retrieval/matching performance.
 In order to differentiate between the two cases we refer to the components of the former as \emph{attack-model}, \emph{attack-resolution}, \emph{attack-FCN}, \emph{attack-pooling} and \emph{attack-whitening} and the latter as \emph{test-model}, \emph{test-resolution}, \emph{test-FCN}, \emph{test-pooling} and \emph{test-whitening}.

\subsection{Image retrieval attacks}
Adversarial attacks for image retrieval are so far limited to the non-targeted case.

\paragraph{Non-targeted mismatch} aims at generating an adversarial image with small perturbation compared to the carrier image and descriptor that is dissimilar to that of the carrier. This is formulated by loss function
\begin{align}
L_{\text{nr}}(\vx_c; \vx) = &  ~\ell_{\text{nr}}(\vx,\vx_c) \hspace{-40pt}&+& \lambda~||\vx - \vx_c||^2\nonumber \\
= &  ~\vh_{\vx}^\top \vh_{\vx_c} &+& \lambda~||\vx - \vx_c||^2.
\label{equ:nontargetretrieval}
\end{align}
The adversarial image is given by minimizer
\begin{equation}
\vx_a = \arg\min_{\vx} L_{\text{nr}}(\vx_c; \vx).
\end{equation}
In this way, the adversary modifies images into their non-indexable counterpart.
The exact formulation in \equ{nontargetretrieval} has not been addressed; the closest is the work of Li \etal~\cite{LJL+18} where they are seeking of a UAP by maximizing $\lone$ descriptor distance instead of minimizing cosine similarity.
\section{Method}
\label{sec:method}
%

We formulate the problem of targeted mismatch attack and then propose various loss functions to address it and to construct concealed query images. 

\begin{figure}
\input{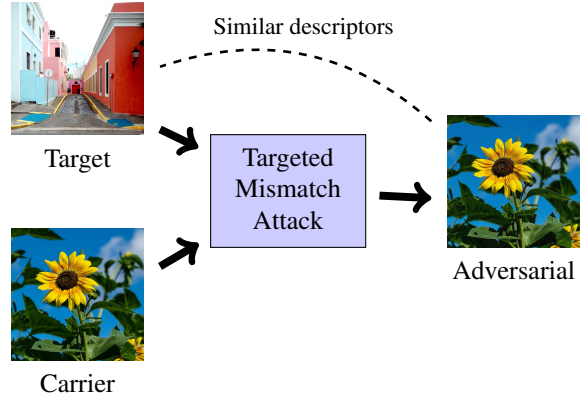}
\caption{In targeted mismatch attacks an adversarial image is generated given a carrier and a target image.
The adversarial image should match the descriptor of the target image but be visually dissimilar to the target; visual dissimilarity to the target is achieved via visual similarity to the carrier.
The attack is formed by a retrieval query using the adversarial image, where the goal is to obtain identical results as with the target query while keeping the target image private. 
\label{fig:tma}}
\end{figure}

\subsection{Problem formulation}
\label{sec:problem}
The adversary tries to generate an adversarial image with the goal of using it as a (concealed) query for image retrieval instead of a \emph{target image}.
The goal is to obtain the same retrieval results without disclosing any information about the target image itself.

We assume a target image $\vx_t \in \real^{W \times H \times 3}$ and a carrier image $\vx_c$ with the same resolution (see Figure~\ref{fig:tma}).
The goal of the adversary is to generate an adversarial image $\vx_a$ that has high \emph{descriptor similarity} but very low \emph{visual similarity} to the target.
Visual (human) dissimilarity is not straightforward to model; we  model visual similarity \wrt another image, \ie the carrier, instead.
We refer to this problem as \emph{targeted mismatch attack} and the corresponding loss function is given by 
\begin{equation}
L_{\text{tr}}(\vx_c, \vx_t; \vx) = \ell_{\text{tr}}(\vx, \vx_t) + \lambda~||\vx - \vx_c||^2.
\label{equ:targetretrieval}
\end{equation}
In Section~\ref{sec:attack} we propose different instantiations of the \emph{performance loss} $\ell_{\text{tr}}$ according to the known and unknown components of the test-model.



\subsection{Targeted mismatch attacks}
\label{sec:attack}
%

In all the following, we assume a white-box access to the FCN, while the whitening is assumed unknown and is totally ignored during the optimization of the adversarial image; its impact on the attack is evaluated by adding it to the test-model.
In general, if all the parameters of the test-model are known, the task is to generate an adversarial image that reproduces the descriptor of the target image. Then, nearest neighbor search will retrieve identical results as if querying with the target image. Choosing a different performance loss introduces invariance or robustness to some parameters of the attacked retrieval system, when these parameters are unknown.
We list different performance loss functions used to minimize ~\equ{targetretrieval}.

\paragraph{Global descriptor.}
Loss function
\begin{equation}
\ell_{\text{desc}}(\vx,\vx_t) = 1 - \vh_{\vx}^\top \vh_{\vx_t}.
\end{equation}
is suitable when all parameters of the retrieval system are known, including the pooling, and when the image is processed by the neural network at its original resolution. Pooling function $h$ is MAC, SPoC, or GeM in our experiments.

\paragraph{Activation tensor.}%
In this scenario, the output of the FCN should be the same for the adversarial and target image, at the original resolution. This is achieved by minimizing the mean squared difference of the two activation tensors
\begin{equation}
\ell_{\text{tens}}(\vx,\vx_t) = \frac{||\vg_{\vx}-\vg_{\vx_t}||^2}{w\cdot h \cdot d}\mbox{.}
\label{equ:tens}
\end{equation}
Identical tensors guarantee identical descriptors computed on top of these tensors, including those where spatial information is taken into account. This covers all global or regional pooling operations, and even deep local features, \eg DELF~\cite{NAS+17}.
However, our experiments show that preserving the activation tensor may result in transferring the target's visual content on the adversarial image (see Figure~\ref{fig:recon}). Further, the visual appearance of the target image can be partially recovered by inverting~\cite{MV15} the activation tensor of the adversarial image.

\paragraph{Activation histogram.}%
Preserving channel-wise first order statistics of the activation tensor, at the original resolution, is a weaker constraint than preserving the exact activation tensor. It guarantees identical descriptors for all global pooling operations that ignore spatial information.
Activation histogram loss function is defined as
\begin{equation}
\ell_{\text{hist}}(\vx,\vx_t) = \frac{1}{d}\sum_{i=1}^d||u(\vg_\vx,\vb)_i - u(\vg_{\vx_t},\vb)_i||\mbox{,}
\end{equation}
%
where $u(\vg_\vx, \vb)_i$ is the histogram of activations from the $i$-th channel of $\vg_\vx$ and $\vb$ is the vector of histogram bin centers.
Histograms are created with soft assignment by an RBF kernel\footnote{We use $e^{\frac{(x-b)^2}{2\sigma^2}}$, where $\sigma=0.1$, $x$ is a scalar  activation normalized by the maximum activation value of the target, and $b$ is the bin center. We uniformly sample bin centers in [0,1] with step equal to 0.05.}.
Compared with the tensor case, the histogram optimization does not preserve the spatial distribution, is significantly faster, and does not suffer from undesirable disclosure artifacts.

\paragraph{Different image resolution.} 
We require an adversarial image at the original resolution of the target ($W\times H$), which when down-sampled to resolution $s$, it retrieves similar results as the target image down-sampled to the same resolution.
This is achieved by loss function
\begin{equation}
L_{\text{tr}}^s(\vx, \vx_t; \vx) = \ell_{\text{tr}}(\vx^s,\vx_t^s) + \lambda~||\vx - \vx_c||^2 \mbox{,} \label{equ:scale}
\end{equation}
where $\ell_{\text{tr}}$ can be any of the descriptor, tensor, or histogram based performance loss functions. Note that~\equ{scale} is different from~\equ{targetretrieval}, the performance loss is computed from re-sampled images, while the distortion loss is still on the original images.

A common down-sampling method used in CNNs is bi-linear interpolation. We have observed that different implementations of such a layer result in different descriptors. The difference is caused by the presence of high-frequencies in the high-resolution image. The adversarial perturbation tends to be high-frequency, therefore different down-sampling results may significantly alternate the result of attack. In order to reduce the sensitivity to down-sampling, we introduce high-frequency removal by Gaussian blurring in the optimization. Instead of~\equ{scale}, the following loss is used
\begin{equation}
L_{\text{tr}}^{\hat{s}}(\vx, \vx_t; \vx) = \ell_{\text{tr}}(\vx^{\hat{s}},\vx_t^{\hat{s}}) + \lambda~||\vx - \vx_c||^2 \mbox{,} \label{equ:blurscale}
\end{equation}
where $\vx^{\hat{s}}$ is image $\vx$ blurred with Gaussian kernel with $\sigma_b$ and then down-sampled. Our experiments show, that blurring plays an important role when the attack-resolution $s$ does not exactly match the test-resolution $s'$, \ie $s' = s+\Delta$.


\paragraph{Ensembles.} We perform the adversarial optimization for a combination of the aforementioned loss functions by minimizing their sum.
Some examples follow.

\begin{itemize}[label={-},leftmargin=2pt]
\setlength\itemsep{2pt}
\item The test-pooling operation is unknown but there is a set $\cP$ of possible pooling operations. 
Minimization of \equ{targetretrieval} is performed for performance loss
\vspace{-4pt}
\begin{equation}
\ell_{\cP}(\vx,\vx_t) =  \frac{\sum_{p \in \cP } \ell_{p}(\vx,\vx_t)}{|\cP|} \mbox{.}
\vspace{-3pt}
\end{equation}

\item The test-resolution is unknown. Joint optimization for a set $\cS$ of resolutions is performed with
\vspace{-2pt}
\begin{equation}
L_{\text{tr}}^{S}(\vx, \vx_t; \vx) = \frac{\sum_{s \in \cS} \ell_{\text{tr}}(\vx^s,\vx_t^s)}{|S|} + \lambda~||\vx - \vx_c||^2.
\vspace{-2pt}
\end{equation}
Any performance loss $\ell_{\text{tr}}$ is used, with or without blurring.
\vspace{-2pt}
\end{itemize}

\subsection{Optimization}
The optimization is performed with Adam and projected gradient descent is used to apply the box constraints, \ie $\vx \in [0,1]^{W\times H\times 3}$. 
The adversarial image is initialized by the carrier image, while after every update its values are clipped to be in $[0,1]$.
The adversarial image is given by 
\vspace{-4pt}
\begin{equation}
\vx_a = \arg \min_{\vx} L_{\text{tr}}(\vx_c, \vx_t; \vx),
\label{equ:attackimage}
\vspace{-3pt}
\end{equation}
where $L_{\text{tr}}$ can be $L_{\text{desc}}$ (with ``desc'' equal to MAC, SPoC, or GeM), $L_{\cP}$, $L_{\text{hist}}$, or $L_{\text{tens}}$ according to the variant, while the variants with multiple scales are denoted \eg by $L_{\text{hist}}^\cS$ without blur or $L_{\text{hist}}^{\hat{\cS}}$ with blur.

\begin{figure*}
\vspace{-8pt}
\centering
\input{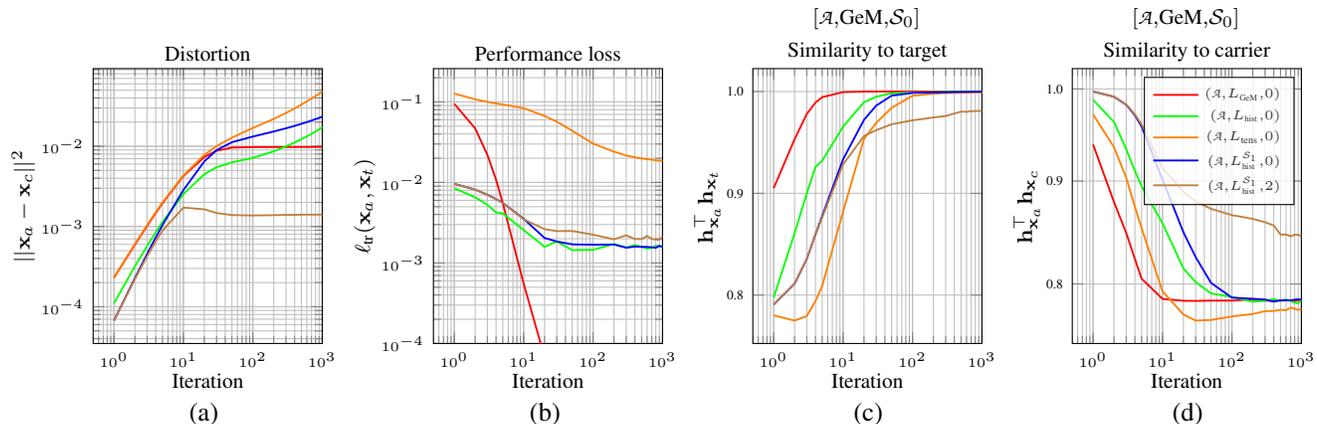}
\begin{tabular}{@{\nmsp}c@{\nmsp}c@{\nmsp}c@{\nmsp}c@{\nmsp}}
& & \footnotesize ~~~~~~~~\test{\alexnet}{GeM}{\scaleorg} & \footnotesize  ~~~~~~~~\test{\alexnet}{GeM}{\scaleorg} \\
{
\begin{tikzpicture}
\begin{axis}[%
	title = {\footnotesize Distortion},
	width=0.265\linewidth,
	height=0.3\linewidth,
	xlabel={\footnotesize Iteration},
	ylabel={\footnotesize $||\vx_a-\vx_c||^2$},
	ymode = log,
	xmode = log,	
	legend cell align={left},
	legend pos=south east,
  legend style={cells={anchor=east}, font =\tiny, fill opacity=0.8, row sep=-2.5pt},
    xmin = 0, xmax = 1000,
    grid=both,
    title style={yshift=-1.5ex,},
  	ylabel shift = -1ex,    
  	xlabel shift = -1ex,    
    ticklabel style = {font=\tiny},
]
\addplot[color=red,solid,line width=0.7] table[x=iter, y expr={\thisrow{gemss}}] \distortalex;
\addplot[color=green,solid,line width=0.7] table[x=iter, y expr={\thisrow{histss}}] \distortalex;
\addplot[color=orange,solid,line width=0.7] table[x=iter, y expr={\thisrow{spatialss}}] \distortalex;
\addplot[color=blue,solid,line width=0.7] table[x=iter, y expr={\thisrow{histms}}] \distortalex;
\addplot[color=brown,solid,line width=0.7] table[x=iter, y expr={\thisrow{histmsdstr2}}] \distortalex;

\end{axis}
\end{tikzpicture}
}

& 

{
\begin{tikzpicture}
\begin{axis}[%
	title = {\footnotesize Performance loss},
	width=0.265\linewidth,
	height=0.3\linewidth,
	xlabel={\footnotesize Iteration},
	ylabel={\footnotesize $\ell_{\text{tr}}(\vx_a,\vx_t)$},
	ymode = log,
	xmode = log,		
	legend cell align={left},
	legend pos=south east,
  legend style={cells={anchor=east}, font =\tiny, fill opacity=0.8, row sep=-2.5pt},
    xmin = 0, xmax = 1000,
    grid=both,
    title style={yshift=-1.5ex,},
  	ylabel shift = -1ex,    
  	xlabel shift = -1ex,    
  	ymin = 0.0001,
    ticklabel style = {font=\tiny},
]
\addplot[color=red,solid,line width=0.7] table[x=iter, y expr={\thisrow{gemss}}] \lossalex;
\addplot[color=green,solid,line width=0.7] table[x=iter, y expr={\thisrow{histss}}] \lossalex;
\addplot[color=orange,solid,line width=0.7] table[x=iter, y expr={\thisrow{spatialss}}] \lossalex;
\addplot[color=blue,solid,line width=0.7] table[x=iter, y expr={\thisrow{histms}}] \lossalex;
\addplot[color=brown,solid,line width=0.7] table[x=iter, y expr={\thisrow{histmsdstr2}}] \lossalex;
\end{axis}
\end{tikzpicture}
}

&

{
\begin{tikzpicture}
\begin{axis}[%
	title = {\footnotesize Similarity to target},
	width=0.265\linewidth,
	height=0.3\linewidth,
	xlabel={\footnotesize Iteration},
	ylabel={\footnotesize $\vh_{\vx_a}^\top \vh_{\vx_t}$},
	xmode = log,	
	legend cell align={left},
	legend pos=south east,
  legend style={cells={anchor=east}, font =\tiny, fill opacity=0.8, row sep=-3pt},
    xmin = 0, xmax = 1000,   	
    grid=both,
    title style={yshift=-1.5ex,},
  	ylabel shift = -1ex,    
  	xlabel shift = -1ex,    
  	ytick = {0.8,0.9,1},
  	yticklabels={0.8,0.9,1.0},
    ticklabel style = {font=\tiny},
]
\addplot[color=red,solid,line width=0.7] table[x=iter, y expr={1-\thisrow{gemss}}] \ipgemtargetalex;
\addplot[color=green,solid,line width=0.7] table[x=iter, y expr={1-\thisrow{histss}}] \ipgemtargetalex;
\addplot[color=orange,solid,line width=0.7] table[x=iter, y expr={1-\thisrow{spatialss}}] \ipgemtargetalex;
\addplot[color=blue,solid,line width=0.7] table[x=iter, y expr={1-\thisrow{histms}}] \ipgemtargetalex;
\addplot[color=brown,solid,line width=0.7] table[x=iter, y expr={1-\thisrow{histmsdstr2}}] \ipgemtargetalex;

\end{axis}
\end{tikzpicture}
}

&

{
\begin{tikzpicture}
\begin{axis}[%
	title = {\footnotesize Similarity to carrier},
	width=0.265\linewidth,
	height=0.3\linewidth,
	xlabel={\footnotesize Iteration},
	ylabel={\footnotesize $\vh_{\vx_a}^\top \vh_{\vx_c}$},
	ymode = log,	
	xmode = log,	
	legend cell align={left},
	legend pos=north east,
  legend style={cells={anchor=east}, font =\tiny, fill opacity=0.8, row sep=-1.2pt},
    xmin = 0, xmax = 1000,
    grid=both,
    title style={yshift=-1.5ex,},
  	ylabel shift = -1ex,    
  	xlabel shift = -1ex,    
  	ytick = {0.8,0.9,1},
  	yticklabels={0.8,0.9,1.0},
    ticklabel style = {font=\tiny},
]
\addplot[color=red,solid,line width=0.7] table[x=iter, y expr={1-\thisrow{gemss}}] \ipgemcarrieralex; \leg{\adve{\alexnet}{\ldesc{GeM}}{0}};
\addplot[color=green,solid,line width=0.7] table[x=iter, y expr={1-\thisrow{histss}}] \ipgemcarrieralex; \leg{\adve{\alexnet}{\lhist}{0}};
\addplot[color=orange,solid,line width=0.7] table[x=iter, y expr={1-\thisrow{spatialss}}] \ipgemcarrieralex; \leg{\adve{\alexnet}{\ltens}{0}};
\addplot[color=blue,solid,line width=0.7] table[x=iter, y expr={1-\thisrow{histms}}] \ipgemcarrieralex; \leg{\adve{\alexnet}{\lhistms{\scalessparse}}{0}};
\addplot[color=brown,solid,line width=0.7] table[x=iter, y expr={1-\thisrow{histmsdstr2}}] \ipgemcarrieralex; \leg{\adve{\alexnet}{\lhistms{\scalessparse}}{2}}
\end{axis}
\end{tikzpicture}
}
\\[-3pt]
~~~~~~~~~(a) & ~~~~~~~~~(b) & ~~~~~~~(c) & ~~~~~~~(d) \\
\end{tabular}
\vspace{-5pt}
\caption{Adversarial images are generated with different loss functions and various measurements are reported as they evolve with the number of iterations. 
The presented measurements are: (a) the distortion \wrt the carrier image, (b) the performance loss from \equ{targetretrieval}, (c) descriptor similarity of the adversarial image to the target for test case \test{\alexnet}{GeM}{\scaleorg} and (d) descriptor similarity of the adversarial image to the carrier for test case \test{\alexnet}{GeM}{\scaleorg}.
The target and carrier images are the ones shown in Figure~\ref{fig:recon}.
\label{fig:iter}
\vspace{-5pt}}
\end{figure*}
\begin{figure}
\vspace{-10pt}
\centering
\input{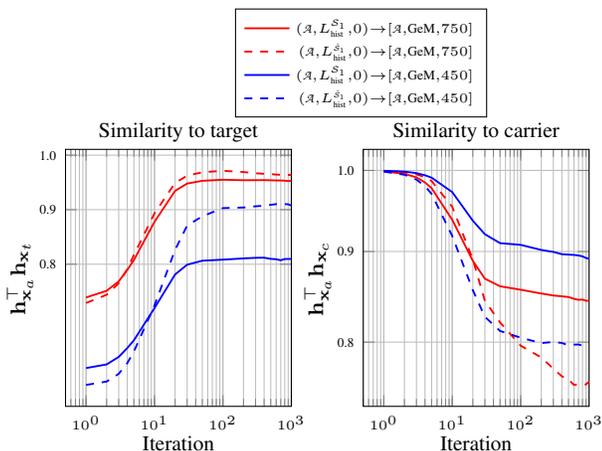}
{
\begin{tikzpicture}
\begin{axis}[%
	title = {\footnotesize Similarity to target},
	width=0.55\columnwidth,
	height=0.6\columnwidth,
	xlabel={\footnotesize Iteration},
	ylabel={\footnotesize $\vh_{\vx_a}^\top \vh_{\vx_t}$},
	xmode = log,	
	legend cell align={left},
	legend pos=south east,
  legend style={cells={anchor=east}, font =\tiny, fill opacity=0.8, row sep=-3pt},
    xmin = 0, xmax = 1000,   	
    grid=both,
    title style={yshift=-1.5ex,},
  	ylabel shift = -1ex,    
  	xlabel shift = -1ex,    
  	ytick = {0.8,0.9,1},
  	yticklabels={0.8,0.9,1.0},
    ticklabel style = {font=\tiny},
]
\addplot[color=red,solid,line width=0.7] table[x=iter, y expr={1-\thisrow{histms750}}] \ipgemtargettransferscale;
\addplot[color=red,dashed,line width=0.7] table[x=iter, y expr={1-\thisrow{histms750b}}] \ipgemtargettransferscale;
\addplot[color=blue,solid,line width=0.7] table[x=iter, y expr={1-\thisrow{histms450}}] \ipgemtargettransferscale;
\addplot[color=blue,dashed,line width=0.7] table[x=iter, y expr={1-\thisrow{histms450b}}] \ipgemtargettransferscale;

\end{axis}
\end{tikzpicture}
}
\hspace{-40pt}
{
\begin{tikzpicture}
\begin{axis}[%
	title = {\footnotesize Similarity to carrier},
	width=0.55\columnwidth,
	height=0.6\columnwidth,
	xlabel={\footnotesize Iteration},
	ylabel={\footnotesize $\vh_{\vx_a}^\top \vh_{\vx_c}$},
	ymode = log,	
	xmode = log,	
	legend cell align={left},
	legend pos=north east,
  legend style={yshift = 12.5ex, xshift = -8ex, cells={anchor=east}, font =\tiny, fill opacity=1, row sep=-1.5pt},
    xmin = 0, xmax = 1000,
    grid=both,
    title style={yshift=-1.5ex,},
  	ylabel shift = -1ex,    
  	xlabel shift = -1ex,    
  	ytick = {0.8,0.9,1},
  	yticklabels={0.8,0.9,1.0},
    ticklabel style = {font=\tiny},
]
\addplot[color=red,solid,line width=0.7] table[x=iter, y expr={1-\thisrow{histms750}}] \ipgemcarriertransferscale;\leg{\trans{\adve{\alexnet}{\lhistms{\scalessparse}}{0}}{\test{\alexnet}{GeM}{750}}};
\addplot[color=red,dashed,line width=0.7] table[x=iter, y expr={1-\thisrow{histms750b}}] \ipgemcarriertransferscale;\leg{\trans{\adve{\alexnet}{\lhistms{\hscalessparse}}{0}}{\test{\alexnet}{GeM}{750}}};
\addplot[color=blue,solid,line width=0.7] table[x=iter, y expr={1-\thisrow{histms450}}] \ipgemcarriertransferscale;\leg{\trans{\adve{\alexnet}{\lhistms{\scalessparse}}{0}}{\test{\alexnet}{GeM}{450}}};
\addplot[color=blue,dashed,line width=0.7] table[x=iter, y expr={1-\thisrow{histms450b}}] \ipgemcarriertransferscale;\leg{\trans{\adve{\alexnet}{\lhistms{\hscalessparse}}{0}}{\test{\alexnet}{GeM}{450}}};

\end{axis}
\end{tikzpicture}
}

\vspace{-3pt}
\caption{Descriptor similarity between the adversarial image and the target or the carrier as it evolves with the number of iterations.
Comparison for test-resolutions that are not in the attack-resolutions for cases without (solid) and with (dashed) blurring.
The adversarial optimization (left) and the test model (right) are denoted with $\rightarrow$.
The target and carrier images are from Figure~\ref{fig:recon}.
\label{fig:itertransfer}
\vspace{-15pt}}
\end{figure}

\section{Experiments}
\label{sec:exp}

Given a test architecture, we validate the success of the targeted mismatch attack in two ways. First, by measuring the cosine similarity between descriptors of the adversarial image $\vx_a$ and the target $\vx_t$ (should be as high as possible), and second, by using $\vx_a$ as an image retrieval query and compare its performance with that of the target query (should be as close as possible)\footnote{Public implementation: \url{https://github.com/gtolias/tma}}.

\subsection{Datasets and evaluation protocol}

We perform experiments on four standard image retrieval benchmarks, namely Holidays~\cite{JDS08}, Copydays~\cite{DJS+09}, \roxf~\cite{RIT+18}, and \rpar~\cite{RIT+18}.
They all consist of a set of query images and a set of database images, while the ground-truth denotes which are the relevant dataset images per query.
We choose to perform attacks only with the first 50 queries for Holidays and Copydays to form adversarial attack benchmarks of reasonable size, while for \roxf and \rpar we keep all 70 of them and use the \emph{Medium} evaluation setup.
All queries are used as targets to form an attack and retrieval performance is measured with mean Average Precision (mAP). 
Unless otherwise stated we use the ``flower'' of Figure~\ref{fig:intro_teaser} as the carrier; it is cropped to match the aspect ratio of the target.
All images are re-sampled to have the largest dimension equal to $1024$, this is the original image resolution.
\roxf and \rpar are treated differently than the other two due to the cropped image queries; the cropped image region that defines the query is used as a target and the relative scale change between queries and database images should be preserved not to affect the ground truth. When the image resolution for descriptor extraction is different than the original one, we down-sample the cropped image with the same scaling factor that the un-cropped one should have been down-sampled with.

\subsection{Implementation details and experimental setup}
We set the learning rate equal to $0.01$ in all our experiments and perform 100 iterations for \ldesc{desc} and \lhist, while 1000 iterations for \ltens.
If there is no convergence, we decrease the learning rate by a factor of 5 and increase the number of iterations by a factor 2 and re-start.
We normalize the distortion term with the dimensionality of $\vx$; this is skipped in the loss function of Sections~\ref{sec:background} and \ref{sec:method}  for brevity.
Moreover, in order to handle the different range of activations for different FCNs, we normalize activation tensors with the maximum target activation before computing the mean squared error in \equ{tens}.
Image blurring at resolution $s$ in \equ{blurscale} is performed by a Gaussian kernel with $\sigma_b= 0.3 \max(W,H) / s$.
The exponent of GeM pooling is always set to 3.

Setting $\lambda=0$ provides a trivial solution to \equ{targetretrieval}, \ie $\vx_a = \vx_t$. 
However, we observe that initialization by $\vx_c$ converges to local minima that are significantly closer to $\vx_c$ than $\vx_t$ even for the case of $\lambda=0$. 
In this way, we satisfy the non-disclosure constraint, \ie the adversarial image is visually dissimilar to the target, and do not sacrifice the performance loss. 
The image distortion \wrt to the carrier image does not sacrifice the goal of concealing the target and preserving user privacy.
Therefore, in our experiments we mostly focus on cases with $\lambda=0$, but also validate cases with $\lambda>0$ to show the impact of the distortion term or in order to promote the non-disclosure constraint for the case of \ltens.

We experiment with different loss functions for targeted mismatch attacks.
We define \scaleorg, \scalessparse, \scalesdense, and \scalesdenselog sets of attack-resolutions\footnote{$\scaleorg= \{1024\}$, $\scalessparse=\scaleorg \cup \{300,400,500,600,700,800,900\}$, $\scalesdense= \scalessparse \cup \{350,450,550,650,750,850,950\}$, $\scalesdenselog = \scaleorg \cup \{262,289,319,351,387,427,470,518,571,630,694,765,843,929\}$}.
We denote AlexNet~\cite{KSH12}, ResNet18~\cite{HZRS16}, and VGG16~\cite{SZ14} by \alexnet, \resnetsmall, and \vgg, respectively.
We use networks that are pre-trained on ImageNet~\cite{RDSK+15} and keep only their fully convolutional part.
The AlexNet and ResNet18 ensemble is denoted by \ensemble; mean loss over two networks is minimized.
We report the triplet attack-model, loss function and value of $\lambda$ to denote the kind of adversarial optimization, for example $\adve{\alexnet}{\lhistms{\scalessparse}}{0}$.
For testing, we report the triplet test-model, test-pooling and test-resolution, for example $\test{\alexnet}{GeM}{\scaleorg}$.

\begin{table}
\newcolumntype{L}[1]{>{\raggedright\let\newline\\\arraybackslash\hspace{0pt}}m{#1}}
\newcolumntype{C}[1]{>{\centering\let\newline\\\arraybackslash\hspace{0pt}}m{#1}}
\newcolumntype{R}[1]{>{\raggedleft\let\newline\\\arraybackslash\hspace{0pt}}m{#1}}
\newcommand\cw{1.3cm}
\def\arraystretch{1.0}
\begin{center}
\setlength{\tabcolsep}{0.0mm}
\scriptsize
\setlength\extrarowheight{1.3pt}
\begin{tabular}{|@{~~~}L{1.2cm}|C{\cw}|C{\cw}|C{\cw}|C{\cw}|C{\cw}|}
    \hline
    \diagbox{$h$}{$L_{\text{tr}}$}  & Original & \ldesc{GeM} & \lpall & \lhist & \ltens \\
    \hline
    \multicolumn{1}{|c|}{} & \multicolumn{1}{c|}{mAP} & \multicolumn{4}{c|}{mAP difference to original} \\
    \cline{2-6}
    GeM   & $41.3$ & $-0.0$ & $-0.0$ & $-0.2$ & $-0.1$ \\
    MAC   & $37.0$ & $-0.5$ & $-0.0$ & $-0.8$ & $-0.0$ \\
    SPoC  & $32.9$ & $-4.4$ & $-0.1$ & $-0.1$ & $-0.7$ \\
    R-MAC & $44.1$ & $-1.2$ & $-0.5$ & $-0.7$ & $-0.0$ \\
    CroW  & $38.2$ & $-1.3$ & $-0.4$ & $-0.2$ & $-0.0$ \\
    \hline
    \multicolumn{1}{|c|}{} &\multicolumn{5}{c|}{$\vx_t^{\scaleto{\top}{1.0ex}} \vx_a$} \\
    \cline{2-6}
    GeM   & $1.000$ & $1.000$ & $1.000$ & $0.997$ & $0.998$ \\
    MAC   & $1.000$ & $0.972$ & $1.000$ & $0.985$ & $0.996$ \\
    SPoC  & $1.000$ & $0.909$ & $1.000$ & $0.999$ & $0.996$ \\
    R-MAC & $1.000$ & $0.972$ & $0.978$ & $0.979$ & $0.997$ \\
    CroW  & $1.000$ & $0.968$ & $0.994$ & $0.995$ & $0.998$ \\
    \hline
\end{tabular}
\end{center}

\caption{Performance evaluation for attacks based on AlexNet and various loss functions optimized at the original image resolution \scaleorg.
Testing is performed on \test{\alexnet}{desc}{\scaleorg} for multiple types of descriptor/pooling.
Mean average Precision on \rpar and mean descriptor similarity between the adversarial image and the target across all queries is reported.
\emph{Original} corresponds to queries without attack.
\label{tab:exp_ss}
}
\end{table}
\begin{figure*}
\vspace{-8pt}
\centering
\input{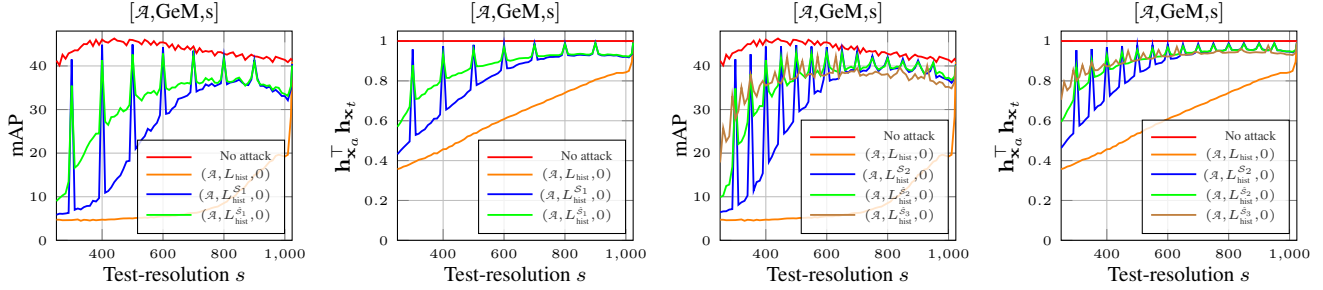}
\begin{tabular}{@{\ssp}c@{\ssp}c@{\ssp}c@{\ssp}c@{\ssp}}

\begin{tikzpicture}
\begin{axis}[%
	title = {\footnotesize \test{\alexnet}{GeM}{$s$}},
	width=0.27\linewidth,
	height=0.25\linewidth,
	xlabel={\footnotesize Test-resolution $s$},
	ylabel={\footnotesize mAP},
	legend cell align={left},
	legend pos=south east,
    legend style={cells={anchor=east}, font =\tiny, fill opacity=0.8, row sep=-2.5pt},
    xmin = 250, xmax = 1024,
    ymin = 0, ymax = 48,
    grid=both,
    title style={yshift=-1.5ex,},
  	ylabel shift = -1ex,    
  	xlabel shift = -1ex,    
    ticklabel style = {font=\tiny},
]
\addplot[color=red,solid,line width=0.7] table[x=imresize, y expr={\thisrow{mapnoattack}}] \expms; \leg{No attack};
\addplot[color=orange,solid,line width=0.7] table[x=imresize, y expr={\thisrow{mapS0}}] \expms; \leg{\adve{\alexnet}{\lhist}{0}};
\addplot[color=blue,solid,line width=0.7] table[x=imresize, y expr={\thisrow{mapS1}}] \expms; \leg{\adve{\alexnet}{\lhistms{\scalessparse}}{0}};
\addplot[color=green,solid,line width=0.7] table[x=imresize, y expr={\thisrow{mapS1blur}}] \expms; \leg{\adve{\alexnet}{\lhistms{\hscalessparse}}{0}};
\end{axis}
\end{tikzpicture}

& 

\begin{tikzpicture}
\begin{axis}[%
	title = {\footnotesize \test{\alexnet}{GeM}{$s$}},
	width=0.27\linewidth,
	height=0.25\linewidth,
	xlabel={\footnotesize Test-resolution $s$},
	ylabel={\footnotesize $\vh_{\vx_a}^\top \vh_{\vx_t}$},
	legend cell align={left},
	legend pos=south east,
    legend style={cells={anchor=east}, font =\tiny, fill opacity=0.8, row sep=-2.5pt},
    xmin = 250, xmax = 1024,
    ymin = 0, ymax = 1.05,
    grid=both,
    title style={yshift=-1.5ex,},
  	ylabel shift = -1ex,    
  	xlabel shift = -1ex,    
    ticklabel style = {font=\tiny},
]
\addplot[color=red,solid,line width=0.7] table[x=imresize, y expr={\thisrow{ipnoattack}}] \expms; \leg{No attack};
\addplot[color=orange,solid,line width=0.7] table[x=imresize, y expr={\thisrow{ipS0}}] \expms; \leg{\adve{\alexnet}{\lhist}{0}};
\addplot[color=blue,solid,line width=0.7] table[x=imresize, y expr={\thisrow{ipS1}}] \expms; \leg{\adve{\alexnet}{\lhistms{\scalessparse}}{0}};
\addplot[color=green,solid,line width=0.7] table[x=imresize, y expr={\thisrow{ipS1blur}}] \expms; \leg{\adve{\alexnet}{\lhistms{\hscalessparse}}{0}};
\end{axis}
\end{tikzpicture}

&

\begin{tikzpicture}
\begin{axis}[%
	title = {\footnotesize \test{\alexnet}{GeM}{$s$}},
	width=0.27\linewidth,
	height=0.25\linewidth,
	xlabel={\footnotesize Test-resolution $s$},
	ylabel={\footnotesize mAP},
	legend cell align={left},
	legend pos=south east,
    legend style={cells={anchor=east}, font =\tiny, fill opacity=0.8, row sep=-2.5pt},
    xmin = 250, xmax = 1024,
    ymin = 0, ymax = 48,
    grid=both,
    title style={yshift=-1.5ex,},
  	ylabel shift = -1ex,    
  	xlabel shift = -1ex,    
    ticklabel style = {font=\tiny},
]
\addplot[color=red,solid,line width=0.7] table[x=imresize, y expr={\thisrow{mapnoattack}}] \expms;\leg{No attack};
\addplot[color=orange,solid,line width=0.7] table[x=imresize, y expr={\thisrow{mapS0}}] \expms; \leg{\adve{\alexnet}{\lhist}{0}};
\addplot[color=blue,solid,line width=0.7] table[x=imresize, y expr={\thisrow{mapS2}}] \expms;\leg{\adve{\alexnet}{\lhistms{\scalesdense}}{0}};
\addplot[color=green,solid,line width=0.7] table[x=imresize, y expr={\thisrow{mapS2blur}}] \expms;\leg{\adve{\alexnet}{\lhistms{\hscalesdense}}{0}};
\addplot[color=brown,solid,line width=0.7] table[x=imresize, y expr={\thisrow{mapS3blur}}] \expms;\leg{\adve{\alexnet}{\lhistms{\hscalesdenselog}}{0}};
\end{axis}
\end{tikzpicture}

& 

\begin{tikzpicture}
\begin{axis}[%
	title = {\footnotesize \test{\alexnet}{GeM}{$s$}},
	width=0.27\linewidth,
	height=0.25\linewidth,
	xlabel={\footnotesize Test-resolution $s$},
	ylabel={\footnotesize $\vh_{\vx_a}^\top \vh_{\vx_t}$},
	legend cell align={left},
	legend pos=south east,
    legend style={cells={anchor=east}, font =\tiny, fill opacity=0.8, row sep=-2.5pt},
    xmin = 250, xmax = 1024,
    ymin = 0, ymax = 1.05,
    grid=both,
    title style={yshift=-1.5ex,},
  	ylabel shift = -1ex,    
  	xlabel shift = -1ex,    
    ticklabel style = {font=\tiny},
]
\addplot[color=red,solid,line width=0.7] table[x=imresize, y expr={\thisrow{ipnoattack}}] \expms; \leg{No attack};
\addplot[color=orange,solid,line width=0.7] table[x=imresize, y expr={\thisrow{ipS0}}] \expms; \leg{\adve{\alexnet}{\lhist}{0}};
\addplot[color=blue,solid,line width=0.7] table[x=imresize, y expr={\thisrow{ipS2}}] \expms; \leg{\adve{\alexnet}{\lhistms{\scalesdense}}{0}};
\addplot[color=green,solid,line width=0.7] table[x=imresize, y expr={\thisrow{ipS2blur}}] \expms; \leg{\adve{\alexnet}{\lhistms{\hscalesdense}}{0}};
\addplot[color=brown,solid,line width=0.7] table[x=imresize, y expr={\thisrow{ipS3blur}}] \expms; \leg{\adve{\alexnet}{\lhistms{\hscalesdenselog}}{0}};
\end{axis}
\end{tikzpicture}

\end{tabular}
\vspace{-5pt}
\caption{
Performance evaluation for attack based on AlexNet and a set of attack-resolutions. 
Mean average Precision on \rpar and mean descriptor similarity between the adversarial image and the target across all queries is shown  for increasing test-resolution.
Comparison using different sets of attack-resolutions and comparison for optimization without ($\cS$) and with ($\hat{\cS}$) image blurring.
\label{fig:exp_ms}
\vspace{-5pt}
}
\end{figure*}
\begin{figure}
\scriptsize
\begin{tabular}{@{\ssp}c@{\ssp}c@{\ssp}c@{\ssp}c@{\ssp}c@{\ssp}}
Target  & Carrier & $\lambda{=}10$ & $\lambda{=}1$ & $\lambda{=}0$ \\[-2pt]
$\vx_t$ & $\vx_c$ & $\vx_a$ & $\vx_a$ &$\vx_a$ \\[0pt]
\includegraphics[height=43pt]{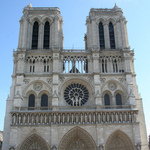}&\includegraphics[height=43pt]{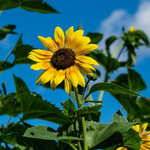}&\includegraphics[height=43pt]{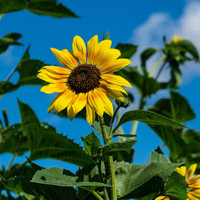}&\includegraphics[height=43pt]{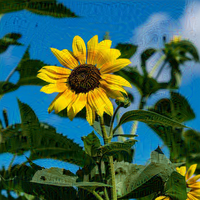}&\includegraphics[height=43pt]{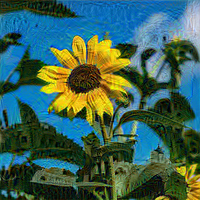}\\[-2pt]
 & 0.702 & 0.873 & 0.987 & 0.998 \\[2pt]
\includegraphics[height=43pt]{fig/distort/notredame.jpg}&\includegraphics[height=43pt]{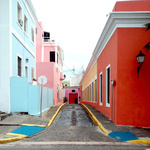}&\includegraphics[height=43pt]{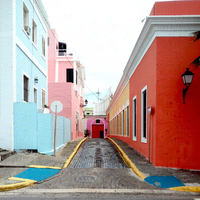}&\includegraphics[height=43pt]{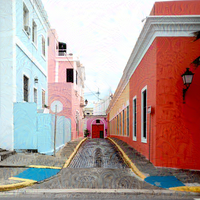}&\includegraphics[height=43pt]{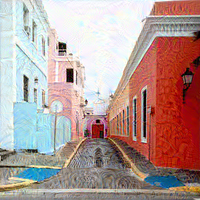}\\[-2pt]
 & 0.796 & 0.953 & 0.995  & 0.999 \\[-2pt]
\end{tabular}
\caption{Adversarial examples for a carrier image and two different targets while optimizing \adve{\alexnet}{\lhistms{\hscalesdense}}{\lambda} for various values of $\lambda$.
Descriptor similarity is reported for $\test{\alexnet}{GeM}{\scaleorg}$.
\label{fig:distort}
\vspace{-15pt}
}
\end{figure}
\begin{figure*}[t]
\vspace{5pt}
\small
\begin{tabular}{@{\ssp}c@{\ssp}c@{\ssp}c@{\ssp}c@{\ssp}c@{\ssp}c@{\ssp}c@{\ssp}c@{\ssp}c@{\ssp}}
Target  & Carrier & \adve{\alexnet}{\ldescms{\tiny GeM}{\hscalessparse}}{0} & \adve{\alexnet}{\lhistms{\hscalessparse}}{0} & \adve{\alexnet}{\lhistms{\hscalessparse}}{1} & \adve{\alexnet}{\ltensms{\hscalessparse}}{0} & \adve{\alexnet}{\ltensms{\hscalessparse}}{1} & \adve{\alexnet}{\ltensms{\hscalesdense}}{0} & \adve{\alexnet}{\ltensms{\hscalessparse}}{1}\\
$\vx_t$ & $\vx_c$ & $\vx_a$ & $\vx_a$ & $\vx_a$ &$\vx_a$ &$\vx_a$ &$\vx_a$ &$\vx_a$ \\ 
\includegraphics[height=70pt]{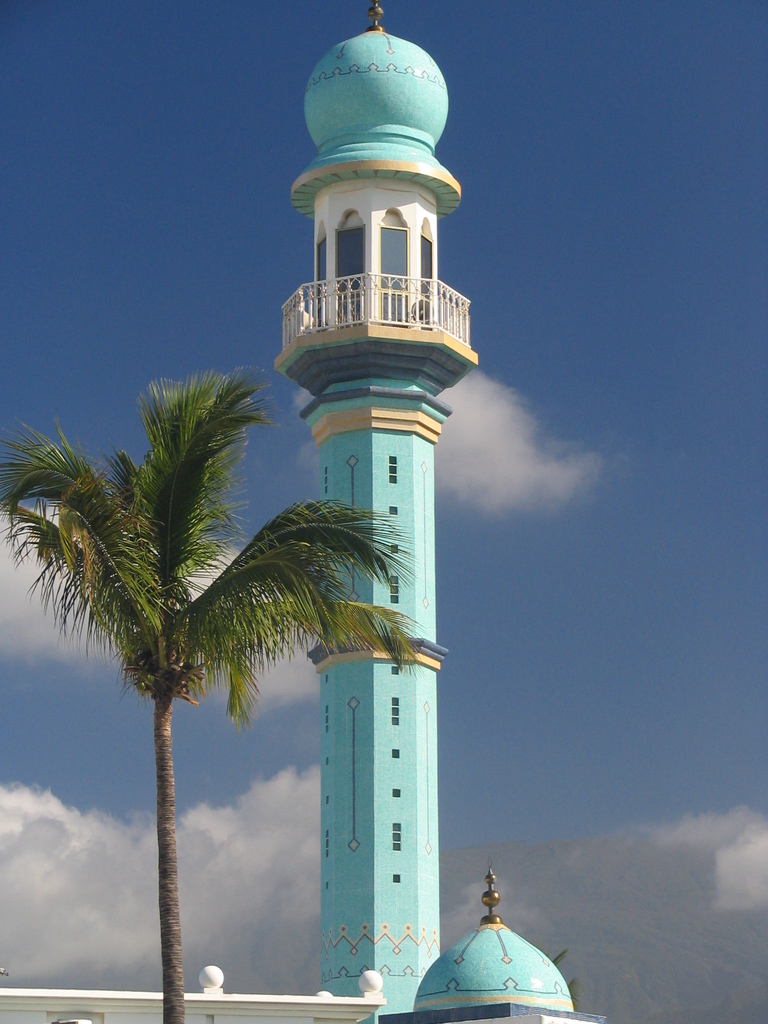}&\includegraphics[height=70pt]{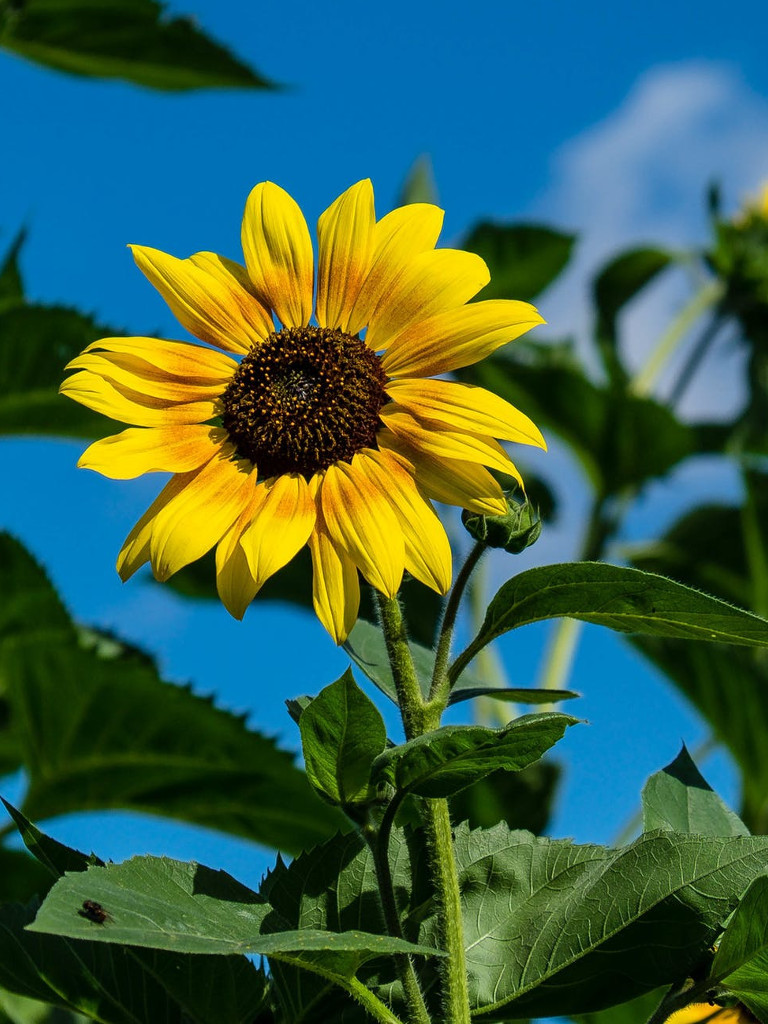}&\includegraphics[height=70pt]{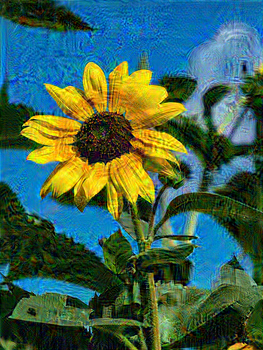}&\includegraphics[height=70pt]{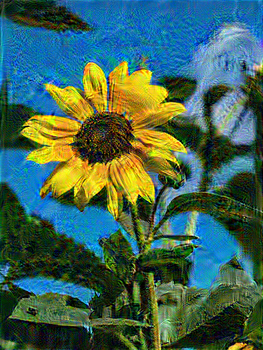}&\includegraphics[height=70pt]{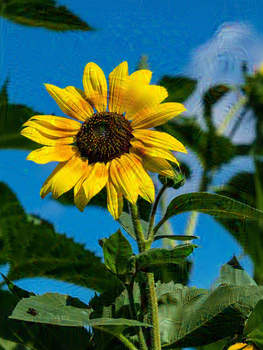}&\includegraphics[height=70pt]{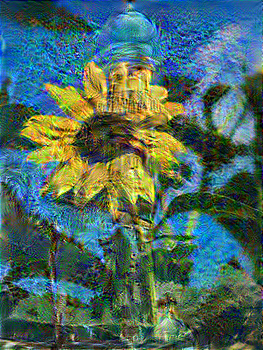}&\includegraphics[height=70pt]{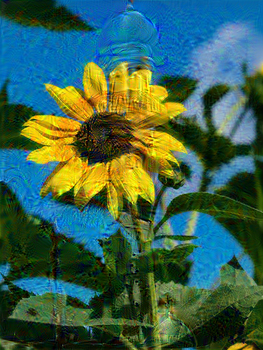}&\includegraphics[height=70pt]{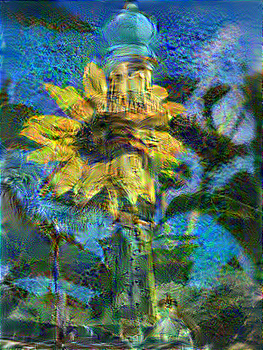}&\includegraphics[height=70pt]{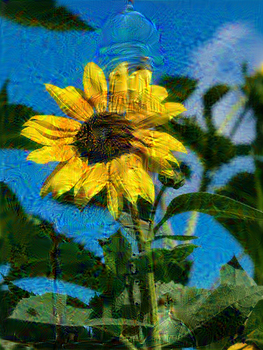}\\
 & 0.782 & 1.000 & 1.000 & 0.994 & 0.999 & 0.997 & 0.998 & 0.997 \\
\includegraphics[height=70pt]{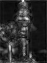}&\includegraphics[height=70pt]{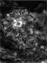}&\includegraphics[height=70pt]{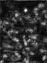} &\includegraphics[height=70pt]{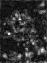}&\includegraphics[height=70pt]{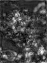}&\includegraphics[height=70pt]{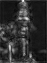}&\includegraphics[height=70pt]{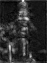}&\includegraphics[height=70pt]{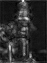}&\includegraphics[height=70pt]{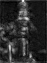}\\[4pt]
\includegraphics[height=70pt]{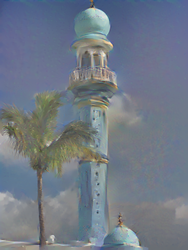}  & \includegraphics[height=70pt]{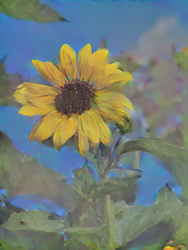} &\includegraphics[height=70pt]{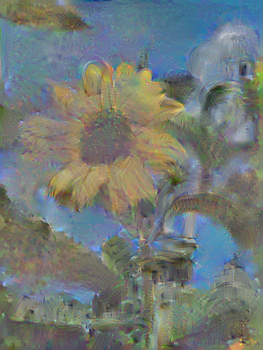} &\includegraphics[height=70pt]{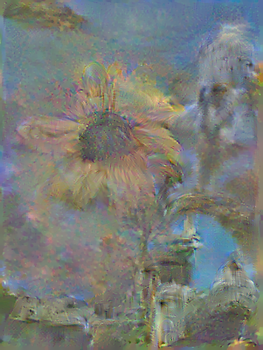}&\includegraphics[height=70pt]{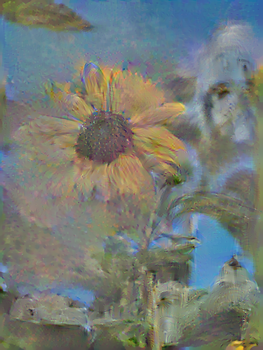}&\includegraphics[height=70pt]{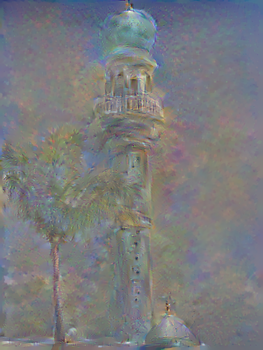}&\includegraphics[height=70pt]{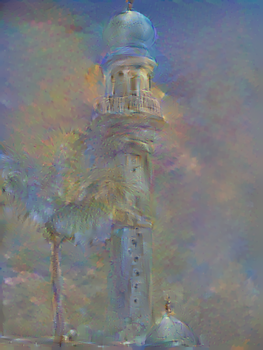}&\includegraphics[height=70pt]{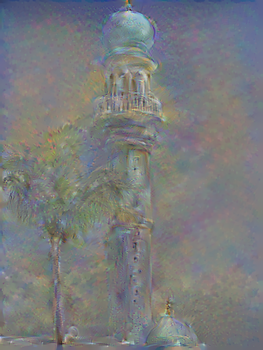}&\includegraphics[height=70pt]{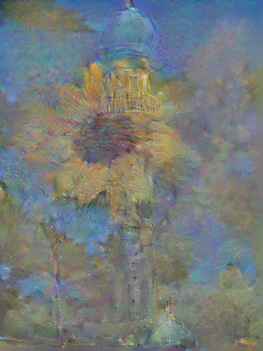} \\
\scalessparse & \scalessparse & \scalessparse & \scalessparse & \scalessparse & \scalessparse & \scalessparse & \scalesdense & $\scalesdense \setminus \scalessparse$ \\
\end{tabular}

\vspace{8pt}
\caption{
Target, carrier and adversarial images for different variants (top image row), a summary of tensor $\vg_\vx$ by depth-wise maximum (middle image row) and the inversion of $\vg_{\vx_t}$, $\vg_{\vx_c}$, or $\vg_{\vx_a}$, respectively, over multiple resolutions (bottom image row). The resolutions for inversion are reported below the bottom row. The tensor inversion shows whether the target, or any information about it, can be reconstructed from the adversarial image. The first two inversions are presented as a reference. Descriptor similarity to the target is reported below the first image row for $\test{\alexnet}{GeM}{1024}$.
\label{fig:recon}
}
\end{figure*}
\subsection{Results}
For each adversarial image we perform the following measurements.
We compute its similarity to the target and to the carrier by cosine similarity of the corresponding descriptors,
we measure the carrier distortion and, lastly, we perform an attack by using it as a query and measure the average precision which is compared to that of the target.

\paragraph{Optimization iterations.} 
We perform the optimization for different loss functions and report the measurements over iteration in Figure~\ref{fig:iter}.
Optimizing global descriptor or histogram converges much faster than the tensor case and results in significantly lower distortion. 
This justifies our choice of using a lower number of iterations for the two approaches.
Increasing the value of $\lambda$ keeps the distortion lower but sacrifices the performance loss, as expected.

In Figure~\ref{fig:itertransfer} we show how the similarity to the target and the carrier evolves for test-resolution that is not included in the set of attack-resolutions. Processing the images with image blurring offers significant improvements, especially for the smaller resolutions.

\paragraph{Robustness to unknown test-pooling.}
In Table~\ref{tab:exp_ss} we present the evaluation comparison for different loss functions and test-pooling. 
The case of same attack-resolution and test-resolution is examined first.
If the test-pooling is directly optimized ($\ldesc{GeM}$ or $\lpall$ case), then perfect performance is achieved.
The histogram and tensor based approaches both perform well for a variety of test-descriptors.

\paragraph{Robustness to unknown test-resolution.}
Cases with different attack-resolution and test-resolution are evaluated and results are presented in Figure~\ref{fig:exp_ms}.
Resolutions that are not part of the attack-resolutions suffer from significant drop in performance when blurring is not performed, while blurring  improves it. We observe how the retrieval performance and descriptor similarity between adversarial image and target are correlated.
Moreover, the optimization for multiple resolutions is clearly better than that for single resolution, while logarithmic sampling of attack-resolutions (\scalesdenselog) significantly  improves the performance for very small test-resolution but harms it for larger ones.

\paragraph{Impact of the distortion term.}
We evaluate \test{\alexnet}{GeM}{\scaleorg} on queries of \rpar for \adve{\alexnet}{\lhistms{\hscalesdense}}{\lambda} and $\lambda$ equal to 0, 0.1, 1, 10. The average similarity between the adversarial image and the target is 0.990, 0.987, 0.956, and 0.767, respectively, while the average distortion is 0.0177, 0.0083, 0.0026, and 0.0008, respectively. Examples of adversarial images are shown in Figure~\ref{fig:distort}. 

\paragraph{Impact of the whitening in the test-model.}
We now consider the case that the test-model includes descriptor whitening. 
The whitening is unknown during the time of the adversarial optimization.
We evaluate the performance on \rpar while learning whitening with PCA on \roxf.
Testing without whitening and \test{\alexnet}{GeM}{\cS_0} or \test{\alexnet}{GeM}{768} achieves 41.3, and 40.2 mAP, respectively.
After applying whitening the respective performances increase to 47.5 and 48.0 mAP.
Attacks with \adve{\alexnet}{\lhistms{\hscalesdense}}{0} achieve 40.2, and 39.4  mAP when tested in the aforementioned cases without whitening.
Attacks with \adve{\alexnet}{\lhistms{\hscalesdense}}{0} achieve 47.3, and 42.9 mAP when tested in the aforementioned cases with whitening. 
Whitening introduces additional challenges, but the attacks seem effective with slightly reduced performance.

\paragraph{Concealing/revealing the target.}
We generate adversarial images for different loss functions and show examples in Figure~\ref{fig:recon}. 
The corresponding tensors show that spatial information is only preserved in the tensor-based loss function. 
The tensor-based approach requires the distortion term to avoid revealing visual structures of the target (adversarial images in 6-th and 7-th column). 
We now pose the question ``can the FCN activations of the adversarial image reveal the content of the target?''. 
To answer, we invert tensor $\vg_{\vx_a}$ at multiple resolutions using the method of Mahendran and Vedaldi~\cite{MV15}. 
The tensor-based approach indeed reveals the target's content in the reconstruction, while no other approach does.
This highlights the benefits of the proposed histogram-based optimization.
Note that the reconstructed image resembles the target less if the resolutions used in the reconstruction are not the same as the attack-resolutions (rightmost column).

\paragraph{Timings.} We report the average optimization time per target image on Holidays dataset  and on a single GPU (Tesla P100) for some indicative cases. Optimizing \adve{\alexnet}{\ldesc{GeM}}{0}, \adve{\alexnet}{\ldescms{GeM}{\hscalessparse}}{0}, \adve{\alexnet}{\lhistms{\hscalessparse}}{0}, \adve{\alexnet}{\lhistms{\hscalesdense}}{0}, and \adve{\alexnet}{\ltensms{\hscalessparse}}{0} takes 1.9, 7.5, 12.3, 22.9, and 68.4 seconds, respectively. Using ResNet18 \adve{\resnetsmall}{\ldesc{GeM}}{0} and \adve{\resnetsmall}{\lhistms{\hscalesdense}}{0} take 3.9 and 40.6 seconds, respectively.

\paragraph{Multiple attacks.}
We show results of multiple attacks in Table~\ref{tab:exp_final_map}. 
We present the original retrieval performance together with the difference in the performance caused by the attack. 
It summarizes the robustness of the histogram and tensor based optimization to unknown pooling operations. It emphasizes the challenges of unknown test-resolution and the impact of the blurring; this outcome can be useful in various different attack models.
The very last row suggests that transferring attacks to different FCNs (optimizing on \ensemble, which includes \alexnet~ and \resnetsmall, and testing on \vgg) is hard to achieve; it is harder than for classification~\cite{SZS+14}.

\begin{table}[t!]
\vspace{2pt}
\newcolumntype{L}[1]{>{\raggedright\let\newline\\\arraybackslash\hspace{0pt}}m{#1}}
\newcolumntype{C}[1]{>{\centering\let\newline\\\arraybackslash\hspace{0pt}}m{#1}}
\newcolumntype{R}[1]{>{\raggedleft\let\newline\\\arraybackslash\hspace{0pt}}m{#1}}
\newcommand\cw{0.5cm}
\newcommand\ccw{0.515cm}
\def\arraystretch{1.3}
\begin{center}
\setlength{\tabcolsep}{0.0mm}
\scriptsize
\setlength\extrarowheight{0pt}
\begin{tabular}{|@{~}L{1.4cm}|@{~}L{1.6cm}|R{\cw}@{ ~/}R{\ccw}@{~}|R{\cw}@{ ~/}R{\ccw}@{~}|R{\cw}@{ ~/}R{\ccw}@{~}|R{\cw}@{ ~/}R{\ccw}@{~}|}
    \hline
    Attack & Test & \multicolumn{2}{c|}{\roxf} & \multicolumn{2}{c|}{\rpar} & \multicolumn{2}{c|}{Holidays} & \multicolumn{2}{c|}{Copydays} \\
    \hline
    \adve{\alexnet}{\lhistms{\hscalesdense}}{0} & \test{\alexnet}{GeM}{\scaleorg} & 
    26.9 & +0.2  & 41.3 & -1.2  & 81.5 & +0.2  & 80.4 & -0.4   \\
    \hline
    \multirow{3}{*}{\adve{\resnetsmall}{\ldescms{\tiny GeM}{\hscalesdense}}{0}} & \test{\resnetsmall}{GeM}{\scaleorg} & 
    21.5 & -0.7  & 46.9 & -0.4  & 82.9 & -0.3  & 69.3 & -0.7   \\
    & \test{\resnetsmall}{GeM}{768} & 
    24.0 & -2.5  & 48.0 & -3.9  & 81.7 & -4.4  & 75.6 & -2.8   \\
    & \test{\resnetsmall}{GeM}{512} & 
    22.4 & -6.7  & 49.7 & -11.1 & 82.8 & -0.6  & 82.1 & -10.7   \\
    \hline
    \multirow{3}{*}{\adve{\resnetsmall}{\lhistms{\scalesdense}}{0}} & \test{\resnetsmall}{GeM}{\scaleorg} & 
    21.5 & -1.2  & 46.9 & -1.9  & 82.9 & -0.6  & 69.3 & -1.3   \\
    & \test{\resnetsmall}{GeM}{768} & 
    24.0 & -3.7  & 48.0 & -7.2  & 81.7 & -2.3  & 75.6 & -7.1   \\
    & \test{\resnetsmall}{GeM}{512} & 
    22.4 & -11.2 & 49.7 & -20.7 & 82.8 & -17.1 & 82.1 & -20.6  \\ 
    \hline
    \multirow{3}{*}{\adve{\resnetsmall}{\lhistms{\hscalesdense}}{0}} & \test{\resnetsmall}{GeM}{\scaleorg} & 
    21.5 & -1.4  & 46.9 & -1.8  & 82.9 & -2.4  & 69.3 & -1.3   \\
    & \test{\resnetsmall}{GeM}{768} & 
    24.0 & -5.3  & 48.0 & -6.0  & 81.7 & -1.7  & 75.6 & -4.2   \\
    & \test{\resnetsmall}{GeM}{512} & 
    22.4 & -7.4  & 49.7 & -11.9 & 82.8& -4.9  & 82.1 & -11.3  \\ 
    \hline
    \adve{\resnetsmall}{\ldesccms{\cP}{\hscalesdense}}{0} & \multirow{3}{*}{\test{\resnetsmall}{CroW}{\scaleorg}} & 
    22.0 & -1.1  & 45.0 & -0.5  & 81.0 & +0.9  & 67.0 & -1.6   \\
    \adve{\resnetsmall}{\lhistms{\hscalesdense}}{0} &  & 
    22.0 & -0.3  & 45.0 & -0.8  & 81.0 & +1.3  & 67.0 & -1.0   \\
    \adve{\resnetsmall}{\ltensms{\hscalesdense}}{0} &  & 
    22.0 & -0.7  & 45.0 & -0.0  & 81.0 & -0.6  & 67.0 & -3.0   \\
    \hline
    \multirow{3}{*}{\adve{\ensemble}{\lhistms{\hscalesdense}}{0}} & \test{\alexnet}{GeM}{\scaleorg} & 
    26.9 & -2.3  & 41.3 & -5.5  & 81.5 & -3.1  & 80.4 & -4.9   \\
    & \test{\resnetsmall}{CroW}{\scaleorg} & 
    22.0 & -1.1  & 45.0 & -0.8  & 81.0 & +1.0  & 67.0 & -0.8   \\
    & \test{\vgg}{GeM}{\scaleorg} & 
    38.1 & -34.9 & 54.0 & -47.4 & 85.7 & -72.6 & 80.0 & -72.9  \\ 
    \hline
\end{tabular}
\end{center}

\vspace{2pt}
\caption{
Performance evaluation for multiple attacks, test-models, and datasets.
Mean average Precision over the original queries, together with the mAP difference to the original caused by the attack, is reported.
The parameters of the adversarial optimization during the attack are shown in the leftmost column, while the type of test-model used is shown in the second column.
\label{tab:exp_final_map}
\vspace{-2pt}
}
\end{table}
%

\section{Conclusions}
\label{sec:conclusion}
We have introduced the problem of targeted mismatch attack for image retrieval and address it in order to construct concealed query images instead of the initial intended query.
We show that optimizing the first order statistics is a good way to generate images that result in the desired descriptors
without disclosing the content of the intended query.
We analyze the impact of image re-sampling, which is a natural component of image retrieval systems and reveal the benefits of simple image blurring in the adversarial image optimization.
Finally, we show that transferring attacks to new FCNs are much more challenging than their image classification counterparts.

We focused on concealing the query in a privacy preserving scenario. In a malicious scenario the adversary might try to corrupt the search results by  targeted mismatch attacks on indexed images. This is an interesting direction and an open research problem.

\small{
\head{Acknowledgments}
Work supported by GA\v{C}R grant 19-23165S and OP VVV funded project CZ.02.1.01/0.0/0.0/16\_019/0000765 ``Research Center for Informatics''.
}

{\small
\bibliographystyle{ieee_fullname}
\bibliography{egbib}
}

\end{document}